\documentclass{article}

\usepackage[preprint]{neurips_2024}

\usepackage[utf8]{inputenc} 
\usepackage[T1]{fontenc}    
\usepackage{hyperref}       
\usepackage{url}            
\usepackage{booktabs}       
\usepackage{amsfonts}       
\usepackage{nicefrac}       
\usepackage{microtype}      
\usepackage{xcolor}         
\usepackage{pifont}%
\usepackage{pgfplots}
\newcommand{\cmark}{\ding{51}}%
\newcommand{\xmark}{\ding{55}}%
\usepackage{multirow}
\usepackage{amsmath}
\usepackage{cleveref}
\usepackage{caption}
\usepackage{graphicx}
\usepackage{subcaption}
\usepackage{wrapfig} 

\definecolor{lightgray}{gray}{0.7}

\usepgfplotslibrary{colorbrewer}
\pgfplotsset{compat = 1.15, cycle list/Set1-8} 
\usetikzlibrary{pgfplots.statistics, pgfplots.colorbrewer} 
\usepackage{pgfplotstable}
\usepackage{filecontents}
\usepackage{algorithm}
\usepackage{algpseudocode}

\title{Seeing Clearly, Forgetting Deeply: Revisiting Fine-Tuned Video Generators for Driving Simulation}

\author{
  Chun-Peng Chang$^{1}$\quad Chen-Yu Wang$^{2}$\quad Julian Schmidt$^{3}$ \\ \textbf{Holger Caesar}$^{1}$ \quad \textbf{Alain Pagani}$^{2}$ \\[2mm]
  $^{1}$~Delft University of Technology \quad 
  $^{2}$~German Research Center for Artificial Intelligence \\
   $^3$~Mercedes-Benz \\[2mm]
}

\begin{document}
\maketitle

\begin{abstract}
Recent advancements in video generation have substantially improved visual quality and temporal coherence, making these models increasingly appealing for applications such as autonomous driving, particularly in the context of driving simulation and so-called "world models". In this work, we investigate the effects of existing fine-tuning video generation approaches on structured driving datasets and uncover a potential trade-off: although visual fidelity improves, spatial accuracy in modeling dynamic elements may degrade. We attribute this degradation to a shift in the alignment between visual quality and dynamic understanding objectives. In datasets with diverse scene structures within temporal space, where objects or perspective shift in varied ways, these objectives tend to highly correlated. However, the very regular and repetitive nature of driving scenes allows visual quality to improve by modeling dominant scene motion patterns, without necessarily preserving fine-grained dynamic behavior. As a result, fine-tuning encourages the model to prioritize surface-level realism over dynamic accuracy. To further examine this phenomenon, we show that simple continual learning strategies, such as replay from diverse domains, can offer a balanced alternative by preserving spatial accuracy while maintaining strong visual quality.

\end{abstract}

\section{Introduction}
Video prediction tasks require generative models to anticipate future frames based on visual cues~\cite{oprea2020review}, which may originate from a single image~\cite{ni2023conditional}, a sequence of preceding frames~\cite{wu2024fairy}, or other conditional inputs~\cite{hu2022make,gao2024vista}. Recent advancements in diffusion-based models~\cite{ho2020denoising} have led to significant improvements in the visual fidelity and temporal coherence of generated videos, bringing them closer to real-world quality. These developments have sparked growing interest in applying video prediction to safety-critical domains such as autonomous driving, where such models have the potential to serve as components of learned "world models"~\cite{wang2024driving}, generative systems that simulate plausible future scenarios to support perception, planning.

Existing works on video prediction emphasize controllability, where generation is guided by conditional signals ranging from high-level language commands to low-level ego vehicle states~\cite{gao2024vista, hassan2024gem}. A common approach involves fine-tuning pre-trained models, such as Stable Video Diffusion (SVD)~\cite{blattmann2023stable}, which are initially trained on large-scale, general-purpose video datasets. 

\begin{figure}
  \centering
   \includegraphics[width=1\linewidth]{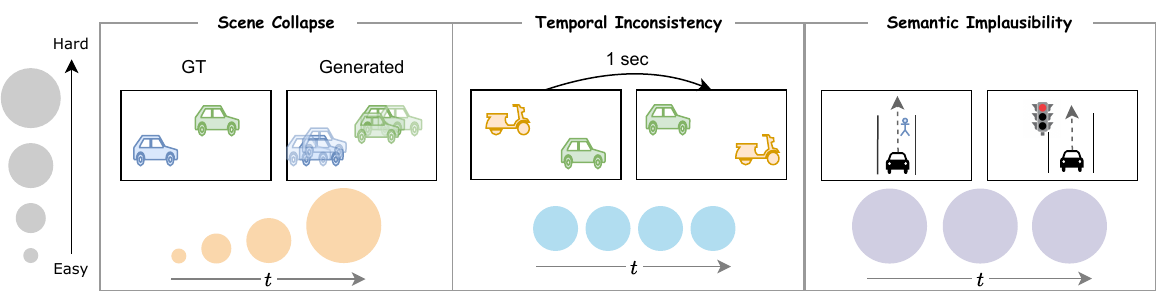}
   \caption{Illustration of common failure types in video generation over time. From left to right: Scene Collapse and Hallucination typically worsen as the rollout progresses; Temporal Inconsistency reflects abrupt changes between nearby frames and can occur at any time; Semantic Implausibility (e.g., ignoring traffic signs) is fundamentally challenging and remains largely unresolved in existing models. The circle sizes indicate relative difficulty, which increases over time or with semantic complexity.}
   \label{fig:teaser}
   \vspace{-5mm}
\end{figure}

However, to serve as a reliable real-world simulator, a video prediction model must not only produce photorealistic and temporally coherent sequences, but also accurately model the motion and spatial behavior of key scene elements, such as vehicles and pedestrians, addressing common failure cases in video generation as illustrated in \Cref{fig:teaser}. Yet, our observations reveal a potential trade-off in existing fine-tuned models: while visual quality improves, the ability to capture scene dynamics does not necessarily benefit, and in some cases, may even degrade, particularly in modeling the relative motion between objects. This suggests that the fine-tuning process can induce a form of catastrophic forgetting~\cite{parisi2019continual,van2019three,wang2024comprehensive,kirkpatrick2017overcoming}, where knowledge acquired during pretraining is diminished as the model adapts to the distribution of the target driving domain.

To better understand this trade-off, we analyze the temporal characteristics of egocentric vision in driving scenes. Traditional egocentric videos, such as those captured by head-mounted cameras~\cite{grauman2022ego4d, pasca2024transfusion,plizzari2024outlook}, typically exhibit diverse temporal scene structure, characterized by frequent shifts in viewpoint, motion patterns, and scene composition. In such settings, improvements in visual fidelity often align with better dynamic modeling, as accurately capturing motion contributes directly to perceptual realism. In contrast, driving videos from datasets like nuScenes~\cite{nuscenes} and OpenDV~\cite{yang2024genad} are highly structured and temporally repetitive, with constrained ego motion and limited changes in perspective. This regularity weakens the natural correlation between visual fidelity and dynamic accuracy, allowing models to enhance appearance by focusing on dominant patterns, such as static layouts or forward motion, without accurately modeling the motion of key objects like vehicles and pedestrians. As a result, fine-tuned models may achieve surface-level realism at the expense of robust motion understanding.

To explore how this forgetting manifests in practice, we examine the effects of fine-tuning across common failure modes in video generation, as illustrated in~\Cref{fig:teaser}. We further show that simple continual learning techniques can help mitigate the degradation in modeling dynamics between objects, providing a viable alternative in scenarios where semantic consistency and dynamic understanding are prioritized alongside visual quality.

In summary, our contributions are as follows:
\begin{itemize}
    \item We reveal a potential trade-off in existing fine-tuned video generators for driving scene prediction, where improvements in visual quality may come at the expense of degraded understanding of scene dynamics.

    \item We conduct a detailed analysis of the gains and losses introduced by fine-tuning, focusing on failure modes such as scene collapse, temporal inconsistency, and semantic implausibility.

    \item We demonstrate that a simple continual learning strategy can mitigate this forgetting problem and serve as a balanced alternative, reducing degradation in dynamic understanding while preserving visual quality. 
\end{itemize}

\section{Related Works}
\noindent{\textbf{Conditional Video Generation: }}
Generating videos based on various conditions has become a popular topic in generative modeling. Common setups include image-to-video~\cite{ni2023conditional,zhao2018learning,hu2022make,ren2024consisti2v,shi2024motion,blattmann2023stable}, video-to-video~\cite{wu2024fairy, liang2024flowvid}, and text-to-video~\cite{hu2022make,chen2024gentron,ho2022video,zhou2024storydiffusion,tian2024videotetris,zhang2023controlvideo} generation. As video generation extends into domains such as autonomous driving, recent works have explored using structured inputs like driving commands or actions to guide the generative process~\cite{wang2024drivedreamer,zhao2024drivedreamer,gao2024vista,hassan2024gem,yang2024genad,hu2023gaia1generativeworldmodel}.
Beyond input conditions, the architectural backbones of video generation models also vary significantly. Earlier approaches often relied on GANs~\cite{arjovsky2017wassersteingan,goodfellow2020generative}; methods like MoCoGAN~\cite{tulyakov2018mocogan} and TGAN~\cite{saito2017temporal} introduced motion-content decomposition and temporal latent variables to synthesize short video clips with basic control over dynamics. More recently, diffusion models~\cite{ho2020denoising} have achieved high-quality and temporally coherent video generation through iterative denoising. Notable examples include Video Diffusion Models~\cite{ho2022video} and SVD~\cite{blattmann2023stable}, which adapt 2D diffusion frameworks with 3D or factorized U-Net architectures to model spatiotemporal dynamics.
In parallel, autoregressive methods have gained traction for their strong capacity to model temporal dependencies through sequential token prediction and Transformer-based designs, such as VideoGPT~\cite{yan2021videogpt}, CogVideo~\cite{hong2022cogvideo}, EmuVideo~\cite{wang2024emu3}, and NOVA~\cite{deng2024autoregressive}.

\noindent{\textbf{Continual Learning: }}
Continual learning, also called lifelong or incremental learning, tackles \emph{catastrophic forgetting}, the degradation of previously acquired knowledge when a model adapts to new data~\cite{parisi2019continual,van2019three,wang2024comprehensive,kirkpatrick2017overcoming}.  
Commonly studied scenarios include \emph{task-incremental} learning, where each task has its own label space and the task identity is provided at test time~\cite{hsu2018re,van2019three}; \emph{class-incremental} learning, where successive tasks introduce novel classes and the model must predict over the union of all classes without task labels~\cite{hsu2018re,van2019three}; and \emph{domain-incremental} learning, in which the label space is fixed but the input distribution shifts, mirroring our transition from generic ego-vision to driving videos~\cite{hsu2018re,van2019three}.  
Strategies for mitigating forgetting fall into several broad families, including \emph{regularisation-based} approaches that penalise parameter or functional drift~\cite{kirkpatrick2017overcoming,zenke2017continual,aljundi2018memory,chaudhry2018riemannian}, \emph{replay-based} schemes that interleave past data through explicit buffers or generative surrogates~\cite{chaudhry2019tiny,riemer2018learning,rebuffi2017icarl,shin2017continual,hou2019learning}, \emph{optimisation-based} methods that reshape gradients or search flatter minima~\cite{wang2021training,javed2019meta,saha2021gradient}, and \emph{representation-based} techniques that promote stable feature spaces via self-supervision or long-horizon pre-training~\cite{madaan2021representational,fini2022self,cha2021co2l}.  
Recent work extends these ideas to the fine-tuning of foundation models~\cite{sun2020ernie, mukhoti2023fine} and explores mixture-of-experts architectures as a capacity-growing mechanism for continual learning~\cite{li2024theory,le2024mixture}.

\noindent{\textbf{Driving Agent Assessment and Scene Simulation: }}
Driving scene simulation is critical for evaluating autonomous agents across perception, prediction, and planning. Traditional open-loop evaluations~\cite{hu2022st, hu2023planning} using pre-recorded videos offer high visual fidelity but lack interactivity, limiting their utility for decision-making assessment~\cite{li2024egostatusneedopenloop}. Simulators such as CARLA~\cite{dosovitskiy2017carla} enable closed-loop testing by responding to agent actions, yet their limited visual realism introduces a domain gap from real-world scenes.
Recent advances in generative models have enabled the construction of photorealistic, interactive environments using both 2D~\cite{hassan2024gem, russell2025gaia, jia2023adriver, lu2024wovogen, gao2024vista, yang2024genad, zhao2024drivedreamer, kim2021drivegan, yang2024drivearena} and 3D~\cite{ljungbergh2024neuroncap, yang2023unisim, lindstrom2024nerfs, zhou2024drivinggaussian, yan2024street, huang2024textit,tian2025drivingforward} representations. These learned world models aim to combine controllability with high-fidelity visual feedback. Despite notable progress, achieving fully realistic and responsive simulation environments remains an open challenge.

\section{Preliminary}
\label{subsec:task}

\subsection{Video Driving-Scene Prediction}
Let $\mathbf{X}_{t-k:t} \in \mathbb{R}^{k \times H \times W \times 3}$ denote the past $k$ RGB frames at time $t$.  
A planner provides a horizon of high-level \emph{conditions} (e.g., action commands, language instructions), denoted by $\mathbf{C}_{t:t+h} = \{\mathbf{c}_{t+1}, \dots, \mathbf{c}_{t+h}\}$ with $\mathbf{c}_{\tau} \in \mathcal{C}$, which the world model $F_{\theta}$ must condition on when forecasting the next $h$ video frames:
\begin{equation}
\hat{\mathbf{X}}_{t+1:t+h} = F_{\theta}(\mathbf{X}_{t-k:t}, \mathbf{C}_{t:t+h}), \quad \hat{\mathbf{X}}_{t+1:t+h} \in \mathbb{R}^{h \times H \times W \times 3}.
\label{eq:prediction}
\end{equation}
The objective is to learn the best model parameters 
\(\theta^\star = \arg\min_{\theta}\; D(\mathbf{X}_{t+1:t+h}, \hat{\mathbf{X}}_{t+1:t+h}),\)
where the distance measure $D(\cdot,\cdot)$ quantifies the similarity between the real and predicted future frames.
However, designing an appropriate distance measure $D$ is itself challenging: there is no single absolute metric that fully captures semantic plausibility, temporal coherence, and visual fidelity of predicted driving scenes. Different choices of $D$ (e.g., pixel-wise losses, perceptual metrics, or learned discriminators) can emphasize different aspects of prediction quality.

Existing approaches like SVD~\cite{blattmann2023stable}, Vista~\cite{gao2024vista}, and GEM~\cite{hassan2024gem} implement \( F_\theta \) using conditional video diffusion models. During training, a clean future video clip \( \mathbf{y}_0 \in \mathbb{R}^{h \times H \times W \times 3} \) is gradually corrupted by Gaussian noise to produce a noisy version \( \mathbf{y}_t \). The model is trained to predict the added noise \( \epsilon \sim \mathcal{N}(0, \mathbf{I}) \) using a denoising network \( \epsilon_\theta(\mathbf{y}_t, t, \mathbf{c}) \), where \( \mathbf{c} \) is the conditioning input. The training objective minimizes the loss \( \mathcal{L}_{\text{train}} = \mathbb{E}_{t, \mathbf{y}_0, \epsilon} \left[ \left\| \epsilon - \epsilon_\theta(\mathbf{y}_t, t, \mathbf{c}) \right\|_2^2 \right] \). At inference time, the generation process starts from pure noise and iteratively applies a denoising step of the form \( \mathbf{y}_{t-1} = f_\theta(\mathbf{y}_t, t, \mathbf{c}) + \sigma_t \mathbf{z} \), where \( \mathbf{z} \sim \mathcal{N}(0, \mathbf{I}) \), until a full video is synthesized. This allows the model to generate temporally coherent future frames conditioned on both past observations and future control signals.

\begin{figure}
  \centering
   \includegraphics[width=1\linewidth]{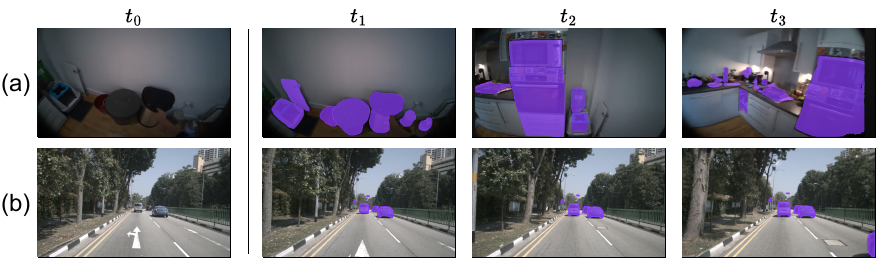}
   \caption{Comparison of semantic-critical actor (SCA) changes in different egocentric vision settings. (a) Egocentric video captured by a camera mounted on a human or robot assistant, where rapid perspective shifts lead to large variations in SCAs over a short time span. (b) Driving scene from a first-person vehicle-mounted camera, showing minimal change in SCAs over the same duration. This highlights the more monotonous and structured nature of SCAs in driving scenarios, making them suitable for studying long-term learning stability.}
   \vspace{-3pt}
   \label{fig:simple_ego}
\end{figure}
\subsection{Semantic Decomposition of Driving Scenes}
Existing video generators~\cite{hassan2024gem, russell2025gaia, jia2023adriver, lu2024wovogen, gao2024vista, yang2024genad, zhao2024drivedreamer, kim2021drivegan, yang2024drivearena} for autonomous driving typically optimize for either \emph{(i)} photorealistic rendering or \emph{(ii)} compliance with control commands such as \textit{turn-left} or \textit{accelerate}. However, an equally important aspect remains under-explored: the faithful modeling of scene elements that are critical for driving decisions. These include pedestrians, traffic signs, and surrounding vehicles, objects that may occupy only a small portion of the frame but are essential for safe planning and control.
We therefore decompose every frame into two disjoint regions:
\begin{enumerate}
    \item \textbf{Semantic-invariant background (SIB)}: large-area elements to which humans are comparatively insensitive (e.g., sky, vegetation, distant buildings, road texture). Surface-level realism is sufficient.
    \item \textbf{Semantic-critical actors (SCA)}: compact yet decisive objects that directly influence driving behaviour (e.g., pedestrians, vehicles, animals). These demand accurate position, identity, and temporal consistency.
\end{enumerate}

Common evaluation metrics also reflect this division. Metrics such as FID~\cite{heusel2017gans} and FVD~\cite{unterthiner2019fvd}, which assess overall visual quality, are often more sensitive to background appearance and global similarity. In contrast, detection or segmentation-based metrics (e.g., mAP, actor segmentation accuracy) focus more directly on the fidelity and consistency of driver-critical actors.

Ideally, the optimal model parameters \(\theta^\star\) would minimize all evaluation measures \(D_1, D_2, \dots, D_k\) simultaneously, achieving strong performance across all aspects of scene prediction. However, in practice, jointly optimizing all \(D_i\) remains extremely difficult. As a result, determining which model is "best" becomes a relatively subjective decision.
Depending on the target application, a scenario-specific optimum can be defined as:
\begin{equation}
\theta^\star_s = \arg\min_{\theta} \sum_{i=1}^k w_i D_i(\mathbf{X}_{t+1:t+h}, \hat{\mathbf{X}}_{t+1:t+h}),
\end{equation}
where \(w_i\) denotes the weight assigned to each metric \(D_i\), reflecting the priorities of a given use case.

Unlike generic video generation tasks, where visual quality and temporal smoothness are often the primary concerns, video generation for driving simulators must also account for driving dynamics and accurate modeling of key actor interactions. At the same time, maintaining a sufficient degree of realism is essential to ensure compatibility with downstream perception and planning systems.

\section{Trade-offs in Fine-tuning for Driving Video }
Fine-tuning general-purpose video generation models for driving tasks introduces both improvements and challenges. In this section, we examine the benefits of enhanced visual quality alongside the unintended degradation in semantic understanding, and explore how the unique properties of driving scenes may contribute to this trade-off.

\subsection{Improved Visual Quality with Trade-offs in Forgetting}
\label{sec:tradeoff}

Recent approaches~\cite{yang2024genad, gao2024vista, hassan2024gem} often adopt a large, general-purpose video generation model such as Stable Video Diffusion (SVD)~\cite{blattmann2023stable} as a base model and fine-tune it for specific tasks like driving simulation. Fine-tuning with domain-specific datasets brings clear benefits~\cite{ding2023parameter}: the model can leverage control signals (e.g., action commands or trajectory plans) to better control the ego-vehicle's viewpoint~\cite{wang2024driving}, and the visual quality of roads, vehicles, and surrounding environments significantly improves thanks to exposure to large-scale driving data.

However, this improvement in surface-level realism comes with a notable trade-off. We observe that fine-tuned models tend to suffer from catastrophic forgetting of higher-level semantic understanding. In particular, behaviors that depend on traffic scene comprehension, such as accurate interaction with semantic-critical actors like pedestrians, traffic lights, and surrounding vehicles, sometimes deteriorate compared to the original pretrained SVD. While the videos appear sharper and more visually consistent, the fine-tuned models often lose subtle decision-critical details essential for realistic driving simulation. See \Cref{sec:experiments} for more experiments and discussion.

\begin{figure}[t]
  \centering

  \begin{minipage}{0.46\linewidth}
    \captionsetup{skip=5pt}
    \centering
    \includegraphics[width=0.9\linewidth]{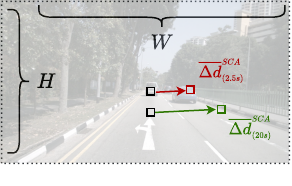}
    \vspace{-3pt}
    \caption{Average displacement of SCAs over 2.5 seconds and 20 seconds relative to image width and height, based on the nuScenes~\cite{nuscenes}. SCAs exhibit limited movement over time.}
    \label{fig:simple_ego_distance}
  \end{minipage}
  \hfill
  \begin{minipage}{0.50\linewidth}
    \centering
    \small
    \captionsetup{type=table} 
    \captionof{table}{Average displacement of different types of SCAs (Semantic Component Areas) over a 2.5-second interval. 
\(\overline{\Delta x}/w_{(2.5s)}\) and \(\overline{\Delta y}/w_{(2.5s)}\) represent the average horizontal and vertical displacements, respectively, normalized by the image width (\(w = 1600\)) and height (\(h = 900\)). 
\(\overline{\Delta d}_{(2.5s)}\) denotes the average total displacement in pixels.}
    \label{tab:next_to_ego_distance}
    \begin{tabular}{c | ccc}
      \toprule
      SCA & $\overline{\Delta x/w}_{(2.5s)} $ & $\overline{\Delta y/w}_{(2.5s)}$ & $\overline{\Delta d}_{(2.5s)}$  \\
      \midrule
      Animal    & 10.5\% & 1.5\% & 173px \\
      Human     & 10.5\% & 0.8\% & 169px \\
      Vehicle   & 10.8\% & 0.9\% & 174px \\
      \hline
      all       & 10.7\% & 0.8\% & 173px \\
      \bottomrule
    \end{tabular}
  \end{minipage}
\end{figure}

\subsection{Diversity of Temporal Scene Structure}
\label{subsec:driving_property}

Here, we investigate the possible causes behind the observed trade-off between improved visual quality and catastrophic forgetting. Conventional ego-vision tasks~\cite{grauman2022ego4d, pasca2024transfusion,plizzari2024outlook} involve rapid viewpoint changes and dynamic scene variations, which pose significant challenges for both understanding and generation compared to standard videos. As a result, existing approaches often pretrain~\cite{pramanick2023egovlpv2} and fine-tune~\cite{pei2024egovideo} models on large-scale datasets, expecting that the model can learn the underlying patterns of complex visual dynamics from large scale of data.

However, as shown in \Cref{fig:simple_ego}, we observe that in driving scenarios, while an advanced understanding of traffic rules is crucial, the visual content associated with semantic-critical actors (SCAs) changes much less than expected despite their crucial role in decision-making. In \Cref{fig:simple_ego_distance} and \Cref{tab:next_to_ego_distance}, we analyze the SCAs in the nuScenes dataset (training/validation and validation splits). On average, each actor moves only 10\% of the width \(W\) and nearly not at all along the height \(H\), suggesting that the motion of SCAs is relatively limited. This visual stability of SCAs suggests a possible cause of the observed trade-off: during fine-tuning, models may prioritize improvements in general visual realism, inadvertently neglecting the subtle, semantically critical behaviors necessary for accurate traffic understanding.


\section{Catastrophic Forgetting for Driving Scene Generation}
\label{sec:forgetting}
\vspace{-3mm}

In the conventional continual learning setting~\cite{parisi2019continual,van2019three,wang2024comprehensive,kirkpatrick2017overcoming}, a model learns from a sequence of tasks \(\mathcal{T}_1, \mathcal{T}_2, \dots, \mathcal{T}_n\), where each task corresponds to a distinct distribution \(P_i(X, Y)\). These tasks may differ in domain or objective . At each time step \(t\), the model receives a dataset \(\boldsymbol{\Gamma}_t = \{(x_t^{(j)}, y_t^{(j)})\}_{j=1}^{N_t}\) for task \(\mathcal{T}_t\), and updates its parameters \(\theta\) to minimize
\(
\mathcal{L}_t(\theta) = \mathbb{E}_{(x, y) \sim \boldsymbol{\Gamma}_t} \left[ \mathcal{L}(f_\theta(x), y) \right].
\)
The challenge arises when optimizing \(\mathcal{L}_t\) leads to degraded performance on prior tasks \(\mathcal{T}_{1}, \dots, \mathcal{T}_{t-1}\).

Unlike traditional catastrophic forgetting, which typically arises from learning across disjoint tasks or domains, the degradation we observe stems from a shift in objective alignment. The observed trade-off arises from a shift in how the objectives of visual quality and scene dynamics align. In general-purpose video prediction, these objectives often reinforce one another, improving visual fidelity typically also enhances temporal and semantic coherence. In such cases, the optimal parameters for visual quality, \(\theta^*_1 = \arg\min_\theta \mathcal{L}_{\text{vis}}(\theta)\), are approximately aligned with those for dynamic accuracy, \(\theta^*_2 = \arg\min_\theta \mathcal{L}_{\text{dyn}}(\theta)\), such that \(\theta^*_1 \approx \theta^*_2\).
However, the relatively structured and monotonous nature of driving datasets weakens this alignment.  In these cases, optimizing for visual fidelity alone leads to \(\theta^*_1\) that no longer approximates \(\theta^*_2\), i.e., \(\theta^*_1 \not\approx \theta^*_2\). This allows the model to improve visual quality by focusing on dominant static patterns and smooth camera motion, even if fine-grained dynamic understanding, such as the motion of pedestrians or scene interactions, is lost. As a result, the gap between these two objectives widens, exposing a fundamental tension during fine-tuning.

\vspace{-3mm}

\section{Experiments}
\label{sec:experiments}
Our experiments are primarily conducted on the nuScenes dataset~\cite{nuscenes}, which provides high-quality, annotated, first-person driving videos with consistent structure and realistic dynamics. We compare the visual quality and object-level positional accuracy between a general-purpose baseline model and a fine-tuned model, as shown in \Cref{tab:model_comparison}, focusing specifically on driving scenarios.
We study three variants of the model. $\mathcal{M}_0$ refers to the baseline model, pre-trained on general video datasets without any domain-specific adaptation. We use the SVD model as the implementation of $\mathcal{M}_0$. The variant $\mathcal{M}_{\mathrm{f.t}}^{\mathrm{r.p.}}$ denotes a fine-tuned version using a simple replay-based strategy, where training samples from the Ego-Exo4D~\cite{grauman2024ego} dataset are intermittently replayed during fine-tuning. This approach helps retain general video understanding while adapting to the specific dynamics and semantics of driving scenes.

\begin{table}
\vspace{-2mm}
  \caption{Comparison of video generation and driving simulation models across various dimensions. The table lists each method’s target task, video resolution, frame rate (FPS), whether ego-centric trajectory and command inputs are considered (Ego Condition), and the underlying model architecture. \\
\textsuperscript{$\ast$}\footnotesize{General ego-centric vision tasks, including but not limited to driving.}}
  \label{tab:model_comparison}
  \scriptsize
  \centering
  \begin{tabular}{l|cccccccccc}
    \toprule
    \multirow{2}{*}{Methods} & \multicolumn{5}{c}{Details} & \multicolumn{2}{c}{Action control} \\
    \cmidrule(r){2-6}
    \cmidrule(r){7-9}
    & Year & Task & Resolution & FPS & Data (hr) & Traj. & Command & Angel\&speed \\
    \midrule
    SVD~\cite{blattmann2023stable} & 2023 & General & \(576 \times 1024\) & 10 & - & \color{red}\xmark & \color{red}\xmark & \color{red}\xmark \\
    Adriver-i~\cite{jia2023adriver} & 2023  & Driving & \(256 \times 512\) & 2 & 300 & \color{red}\xmark & \color{green}\cmark & \color{red}\xmark \\
    GAIA-1~\cite{hu2023gaia1generativeworldmodel} & 2023  & Driving & \(288 \times 512\) & 2 & 4.7k & \color{red}\xmark & \color{red}\xmark & \color{green}\cmark \\
    DriveDreamer~\cite{wang2024drivedreamer} & 2024 & Driving & \(128 \times 192\) & 12 & 5 & \color{red}\xmark & \color{red}\xmark & \color{green}\cmark \\
    Drive-WM~\cite{wang2024driving} & 2024  & Driving & \(192 \times 384\) & 2 & 5 & \color{green}\cmark & \color{red}\xmark & \color{red}\xmark \\
    WoVoGen~\cite{lu2024wovogen} & 2024  & Driving & \(256 \times 448\) & 2 & 5 & \color{red}\xmark & \color{red}\xmark & \color{green}\cmark \\
    GenAD~\cite{yang2024genad} & 2024  & Driving & \(256 \times 448\) & 2 & 2k & \color{green}\cmark & \color{green}\cmark & \color{red}\xmark \\
    Vista~\cite{gao2024vista} & 2024 & Driving & \(576 \times 1024\) & 10 & 1.7k & \color{green}\cmark & \color{green}\cmark & \color{green}\cmark \\
    GEM~\cite{hassan2024gem} & 2024  & Ego-Vis\textsuperscript{$\ast$} & \(576 \times 1024\) & 10 & 4.2k & \color{green}\cmark & \color{green}\cmark  & \color{red}\xmark \\
    GAIA-2~\cite{russell2025gaia} & 2025  & Driving & \(448 \times 960\) & - & 140K & \color{green}\cmark & \color{green}\cmark & \color{green}\cmark \\    
    \bottomrule
  \end{tabular}
  \vspace{-2mm}
\end{table}
\subsection{Evaluation \& Metric}
\label{sec:eval_metric}

Evaluating generated frames in video prediction is challenging and often subjective, as no single metric provides a complete assessment. The choice of evaluation metric depends on the specific aspect being measured. For visual quality, FID~\cite{heusel2017gans} and FVD~\cite{unterthiner2018towards} are commonly used. Following the protocol from Vista~\cite{gao2024vista}, all images are center-cropped and resized to \(256 \times 448\)\footnote{\tiny\url{https://github.com/OpenDriveLab/Vista/blob/main/vwm/data/subsets/common.py\#L34-L51}}, and FVD is computed using 25-frame clips resized to \(224 \times 224\).
To assess object-level fidelity, we leverage SAM2~\cite{ravi2024sam} with ground-truth bounding boxes as prompts to track all annotated SCAs in the nuScenes dataset. The resulting masks are used to analyze both spatial consistency and mask-level shape differences. Our object localization results are consistent with prior methods~\cite{xing2025openemma, hassan2024gem}, which typically rely on YOLO~\cite{khanam2024yolov11}; however, our approach also incorporates mask-based shape analysis for finer-grained evaluation.
In addition, we conduct human evaluation on cases that require high-level semantic understanding, such as interpreting traffic signs, where automatic metrics may fall short.

\subsection{Discussion}
\label{sec:discussion}
\noindent{\bf Does Existing Methods Improve Visual Quality?}

In \Cref{tab:fid_fvd}, we compare the FID and FVD scores of both the baseline model (SVD) and the model fine-tuned on driving scenes. The results suggest that state-of-the-art fine-tuning methods achieve superior performance in terms of visual quality.

\begin{wrapfigure}{r}{0.5\textwidth}
\vspace{-4mm}
    \centering
    \captionsetup{skip=5pt}
    \includegraphics[width=\linewidth]{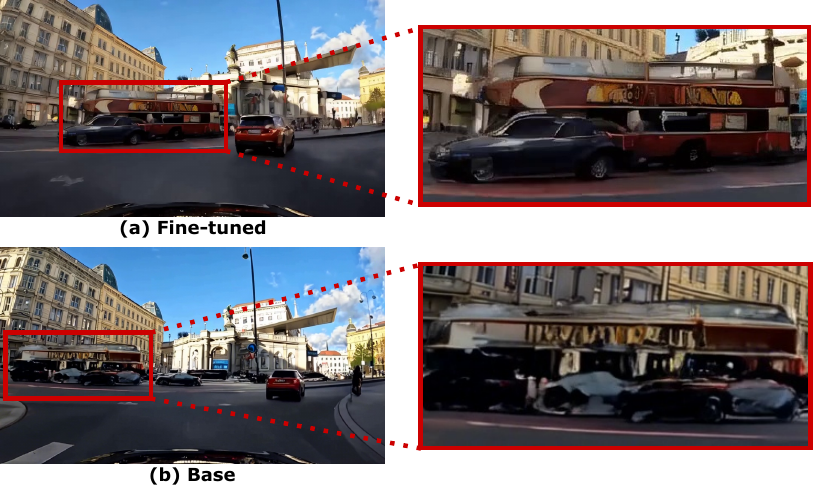}
    \caption{The fine-tuned model demonstrates better visual quality, particularly for moving vehicles. In contrast, the baseline model sometimes produces collapsed objects and unrealistic motion.}
    \label{fig:visual_quality}
    \vspace{-3mm}
\end{wrapfigure}

Given the limitations of FID and FVD~\cite{blattmann2023align,gao2024vista,wu2024towards}, human evaluation remains a widely adopted alternative for assessing visual quality~\cite{gao2024vista,hassan2024gem,chen2024videocrafter2,girdhar2024factorizing,wang2025swap,wang2025lavie}. Results from recent works such as GEM~\cite{hassan2024gem} and Vista~\cite{gao2024vista} show that human evaluators consistently prefer fine-tuned models when judging visual quality. As model-based evaluators gain popularity for subjective tasks, studies have reported high agreement between human judgments and LLM-based assessments~\cite{zheng2023judgingllmasajudgemtbenchchatbot,chiang2023can}. In our experiments, we adopt a GPT-4o evaluator following the Two-Alternative Forced Choice protocol described in Vista, and observe similar results, supporting the conclusion that fine-tuning improves visual quality. See appendix for more details.

\textit{Our answer: Yes, fine-tuning improves visual quality, especially the visual details of vehicle.}

\begin{table}[t]
\centering
\captionof{table}{Comparison of visual quality using FID and FVD scores. The results are computed on 5k samples from the nuScenes validation split. For FVD, each clip contains 25 frames, details can be found in \Cref{sec:eval_metric}.\textcolor{red}{\textsuperscript{$\ast$}}\footnotesize{w/o action control or ego status information.} }
\label{tab:fid_fvd}
\scriptsize
\renewcommand{\arraystretch}{1.4}
\begin{tabular}{l|ccccccc|cc}
\toprule
 Metrics & DriveGAN & DriveDreamer & WoVoGen & Drive-WM & GenAD & GEM & Vista & \tiny{\(\mathcal{M}_{\mathrm{f.t}}^{r.p.}\)}\textcolor{red}{\textsuperscript{$\ast$}} & \tiny{\(\mathcal{M}_{0}\)}\textcolor{red}{\textsuperscript{$\ast$}} \\
\midrule

FID \(\downarrow\) & 73.4 & 52.6 & 27.6 & 15.8 & 15.4 & 10.5 & \bf 6.9 & 16.8 & 20.7 \\
FVD \(\downarrow\) & 502.3 & 452.0 & 417.7 & 122.7 & 184.0 & 158.5 & \bf 89.4  & 124.6 & 128.4 \\

\bottomrule
\end{tabular}
\vspace{-4mm}
\end{table}

\begin{table}[t]
\centering
\caption{
Evaluation of each model's ability to accurately predict how long Semantic-Critical Actors (SCAs) remain present in the generated video.
\textbf{M} (\%): matching rate, percentage of samples where the predicted appearing length matches the ground-truth. 
\textbf{FP} (\%): false positive, indicates the percentage of SCAs that remain in the predicted scene longer than in the ground-truth scene.
\textbf{FN} (\%): false negative, indicates the percentage of SCAs that remain in the predicted scene shorter than in the ground-truth scene..
\textbf{P.} (\%): precision. 
\textbf{R.} (\%): recall. 
\textbf{f.t.}: whether the model is fine-tuned on the same data split. Results are reported separately for training and validation sets.
}

\label{tab:duration_stats}
\scriptsize
\renewcommand{\arraystretch}{1.4}
\begin{tabular}{l|cccccc cccccc}

\toprule
\multirow{2}{*}{Methods} & \multicolumn{6}{c}{nuscene Train} & \multicolumn{6}{c}{ nuscene Val.} \\
\cmidrule(r){2-7}
\cmidrule(r){8-13}
 & M(\%) \(\uparrow\) & FP(\%) \(\downarrow\) & FN(\%) \(\downarrow\) & P. & R. & f.t. & M(\%) \(\uparrow\) & FP(\%) \(\downarrow\) & FN(\%) \(\downarrow\) & P. & R. & f.t. \\
\midrule
\(\mathcal{M}_{\mathrm{f.t}}\) & 60.8 & 24.1 & 15.9 & 71.6 & 79.3 & \color{green}\cmark & 62.8 & 22.9 & 14.3 & 73.3 & 81.5 & \color{red}\xmark \\
\(\mathcal{M}_{\mathrm{f.t}}^{r.p.}\) & 61.2 & 24.3 &  14.3 & 71.7 & 81.1 & \color{green}\cmark & 63.3 & 24.4 & 12.1 & 72.2 & 84.0 & \color{red}\xmark \\
\(\mathcal{M}_{0}\) & 60.9 & 25.9 & 14.1 & 70.2 & 81.2 & \color{red}\xmark & 63.5 & 25.5 & 11.0 & 71.4 & 85.2 & \color{red}\xmark \\
\bottomrule
\end{tabular}
\end{table}

\begin{figure}[t]
  \centering
  \begin{minipage}{0.53\linewidth}
    \captionsetup{skip=5pt}
    \centering
    \includegraphics[width=1\linewidth]{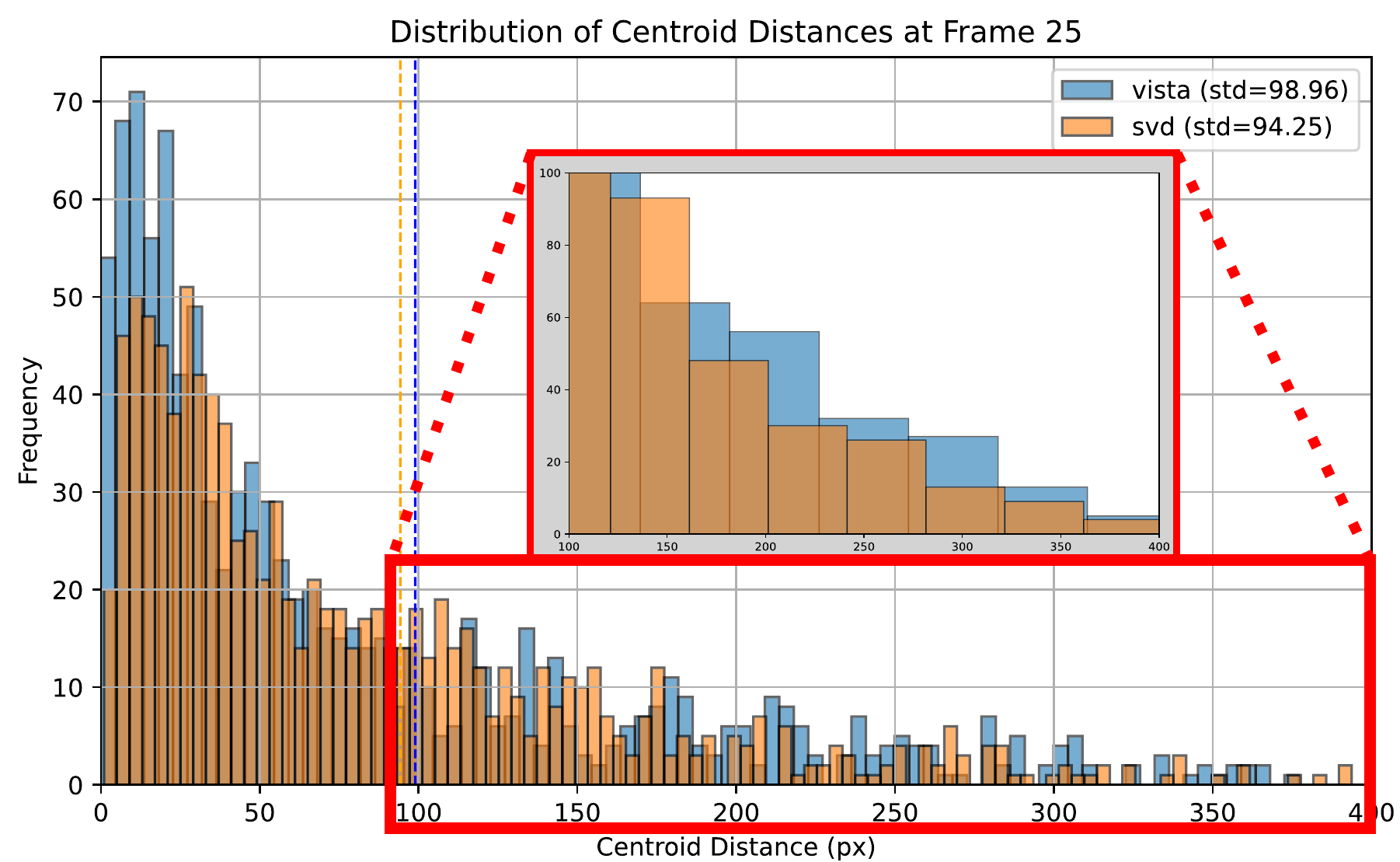}
    \caption{Distribution of centroid distances between predicted and ground-truth object locations at frame 25. The histogram compares the fine-tuned model (Vista) and the baseline model. The fine-tuned model exhibits a higher standard deviation and a longer tail, indicating more frequent large deviations. The zoomed-in region highlights the increased number of high-displacement outliers in the fine-tuned model.}
    \label{fig:box_std}
  \end{minipage}
  \hfill
  \begin{minipage}{0.45\linewidth}
  \captionsetup{skip=5pt}
    \centering
    \begin{tikzpicture}
    \begin{axis}[
        width=\linewidth, 
        height=0.7\linewidth, 
        xlabel={Frame Number},
        ylabel={Accuracy(\%)},
        xmin=1, xmax=24,
        ymin=88, ymax=100, 
        xtick={4, 8, 12, 16, 20, 24},
        ytick={90, 95, 100}, 
        grid=major,
        grid style={solid, draw=gray!50}, 
        ymajorgrids=true, 
        xmajorgrids=false, 
        legend pos=south west,
        legend style={font=\footnotesize},
        tick style={draw=none},
        every axis plot/.append style={thick, mark=none},
        cycle list name=color list,
        ylabel style={yshift=-7pt} 
    ]
    
    \addplot+[purple] coordinates {
        (0, 100) (1,99.9) (2,99.6) (3,99.3) (4,98.7) (5,98.5) (6,98.3) (7,97.4) (8,96.5) (9,95.9) (10,95.7) (11,95.8) (12,95.6) (13,95.1) (14,94.3) (15,93.6) (16,93.4) (17,92.9) (18,91.9) (19,92.0) (20,91.8) (21,91.4) (22,90.4) (23,89.7)(24, 88.9)
    };
    \addlegendentry{\(\mathcal{M}_{\mathrm{f.t}}\)}

    \addplot+[lightgray] coordinates {
        (0, 100) (1,99.6) (2,99.4) (3,99.1) (4,98.8) (5,98.4) (6,97.9) (7,97.5) (8,97.0) (9,96.7) (10,96.3) (11,95.9) (12,95.6) (13,95.1) (14,94.8) (15,94.8) (16,94.4) (17,94.0) (18,93.8) (19,93.5) (20,93.1) (21,92.6) (22,92.4) (23,91.2)(24, 90.9)
    };
    \addlegendentry{\(\mathcal{M}_{\mathrm{f.t}}^{r.p.}\)}
    
    \addplot+[green] coordinates {
        (0, 100) (1,99.5) (2,99.3) (3,98.7) (4,98.2) (5,97.9) (6,97.3) (7,97.0) (8,96.8) (9,96.8) (10,96.4) (11,96.0) (12,95.9) (13,95.9) (14,95.4) (15,95.4) (16,94.8) (17,95.1) (18,94.0) (19,94.1) (20,93.9) (21,93.3) (22,92.7) (23,91.8)(24, 91.2)
    };
    \addlegendentry{\(\mathcal{M}_{0}\)}
    \end{axis}
\end{tikzpicture}
    \caption{ Accuracy of SCAs presence matching over time. The plot shows the proportion of correctly matched SCAs between generated and ground-truth frames. While the fine-tuned model performs comparably in the early frames, its accuracy declines more rapidly over time compared to the baseline model. The initially higher performance of the fine-tuned model is partly due to early scene collapse in the baseline model.}
    \label{tab:line_n}
  \end{minipage}
  \vspace{-5mm}
\end{figure}

\noindent{\bf Does Existing Fine-Tuning Methods Improve Scene Dynamic Understanding of Semantic-Critical Actors?}
For a fair comparison, the following experiments mentioned are conducted without providing additional information, such as ego vehicle status or control signals.
In \Cref{tab:duration_stats}, we analyze how accurately each method predicts the duration of SCA presence in video clips. The results indicate that the fine-tuned model (Vista) does not show improvement across various metrics, including match rate, false positives, and recall. Furthermore, a comparison between the training and validation splits reveals similar scores, even though the fine-tuned model was trained on the training split. This suggests that the model has not effectively learned to capture object duration, as performance does not improve even on seen data.
\Cref{fig:box_std,tab:line_n} provide quantitative evidence of the fine-tuned model's degradation in dynamic scene understanding. \Cref{fig:box_std} shows the distribution of centroid distances between predicted and ground-truth object positions at frame 25, comparing the fine-tuned model (Vista) and the baseline model. The fine-tuned model exhibits a longer tail and higher standard deviation, indicating a greater number of large displacement errors. The zoomed-in region further reveals a noticeable increase in high-displacement outliers, suggesting reduced spatial stability. Complementing this, \Cref{tab:line_n} tracks the accuracy of SCAs by checking whether each object present in the ground-truth video continues to appear in the corresponding frames of the predicted video. While the fine-tuned model initially performs comparably, or even slightly better due to fewer early scene collapses, it exhibits a sharper decline in accuracy over time. This suggests that when it comes to predicting the gradual disappearance of objects from the scene, the baseline model remains more reliable.

\textit{Our answer: No, unlike visual quality, fine-tuning offers limited benefit for dynamic understanding and can sometimes worsen interaction reasoning.}


\noindent{\bf Does the Existing Model Understand Traffic Signs?}
To further verify whether existing models can recognize traffic signs and rules, an essential property for driving simulation, we manually filter challenging scenes from the nuScenes validation split. These include cases such as left/right turn only and red light scenarios, with 200 samples containing 25 frames each (2.5 seconds). Given that evaluation of this task requires rich human knowledge, we rely on human evaluation to determine whether the vehicle in each clip obeys the traffic rules. For each clip, human reviewers are required to choose between "correct" or "incorrect" based on the vehicle’s compliance with the observed traffic rule.
The results suggest that for left/right turn only signs, both models, the baseline model (SVD) and the fine-tuned model (Vistas), correctly generate the corresponding behavior in only 5\% of the cases without explicitly receiving action control signals as input. Notably, the few correct predictions typically occur when the vehicle is already midway through the turn, indicating that the models may be reacting to the ongoing motion rather than understanding the traffic sign itself.
For red light cases, Vista correctly stops the vehicle in 12\% of the clips, while SVD achieves 30\%. However, in scenarios where the vehicle does stop correctly, we often observe that this occurs when many other cars are already stopped in front of the ego vehicle. This suggests that the models are likely relying on visual scene patterns rather than demonstrating genuine understanding of traffic signals.

\textit{Our answer: Likely not. The few successful cases appear to rely on strong visual cues, such as the presence of multiple stopped vehicles or the vehicle already being midway through a turn, rather than a genuine understanding of traffic signs and rules.}

\noindent{\bf Long-Horizon in Driving Scene Simulation}
In our experiments, we primarily focus on short-horizon predictions of up to 2.5 seconds, denoted as $\hat{\mathbf{X}}_{t+1:t+h}$ where $h < 25$, rather than longer rollouts (e.g., 15 seconds) explored in some prior work~\cite{gao2024vista, hassan2024gem}. While long-horizon prediction is an interesting direction and introduces additional challenges~\cite{russell2025gaia, gao2024vista, hassan2024gem}, it also faces two major issues: object collapse and hallucination. These issues are not entirely due to model limitations; in complex urban environments, a large portion of the future scene is inherently unobservable from a single viewpoint. Even experienced human drivers cannot accurately anticipate all future events over such extended time spans. As a result, evaluating model performance, particularly in terms of semantic-critical actors, becomes difficult and less meaningful under long-horizon settings.

\noindent{\bf Are Existing Methods Good Enough for Driving Simulation?}
While existing video generation models demonstrate impressive achievements in controllability and visual quality, we argue that they are still insufficient for realistic driving simulation. Specifically, current methods often fail to capture and understand complex driving dynamics beyond simple motion patterns and surface-level visual fidelity. As a result, they lack the depth of reasoning required to simulate traffic rules, agent interactions, and other high-level semantic behaviors crucial for real-world driving scenarios.


\begin{figure}
\captionsetup{skip=3pt}
  \centering
   \includegraphics[width=0.9\linewidth]{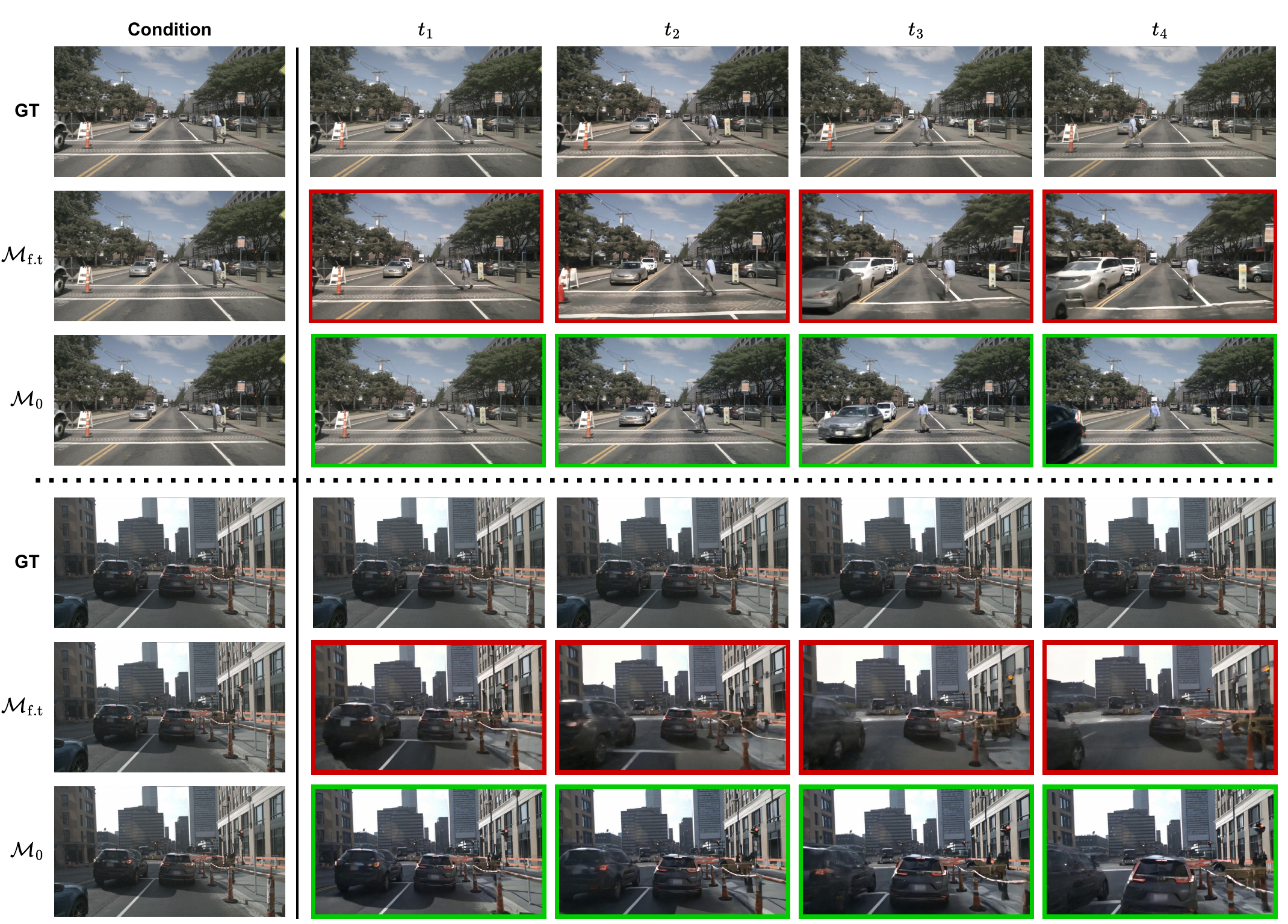}
   \caption{
Illustration of the trade-off between visual quality and high-level semantic understanding after fine-tuning. The pretrained SVD~\cite{blattmann2023stable}, despite its lower visual fidelity, correctly stops the ego vehicle for a crossing pedestrian. In contrast, the fine-tuned model (e.g., Vista~\cite{gao2024vista}) produces sharper visuals but fails to respond to the pedestrian, resulting in unrealistic and unsafe behavior.
}
   \label{fig:forgetting}
   \vspace{-3mm}
\end{figure}
\section{Conclusion}
\vspace{-3mm}
This work investigates the effects of fine-tuning video prediction models for driving scenes and highlights a potential trade-off: while fine-tuning often enhances visual quality, it does not necessarily improve, and can sometimes degrade, the model’s ability to accurately capture scene dynamics. Given the subjective nature of video quality assessment, visual improvements should not be assumed to reflect better semantic understanding.
To support this point, we compare pre-trained and fine-tuned models in terms of their ability to predict the locations of key scene elements such as vehicles and pedestrians. Our findings indicate that existing fine-tuning methods may reduce spatial precision, possibly due to overfitting to the structured patterns of driving datasets. We show that simple continual learning strategies, such as replay from diverse domains, can serve as a balanced alternative, preserving spatial accuracy while maintaining good visual quality. These results suggest that future work in video prediction should consider both appearance and semantic consistency, particularly for safety-critical applications.

\clearpage
{
    \small
    \bibliographystyle{plain}
    \bibliography{main}
}


\newpage
\appendix

\section{Terminology}

In our experiments, we use Vista~\cite{gao2024vista} as a representative fine-tuned model due to its state-of-the-art performance on visual quality metrics, including FID and FVD. Since Vista is fine-tuned from Stable Video Diffusion (SVD)~\cite{blattmann2023stable}, we denote the original SVD model as our baseline \(\mathcal{M}_{0}\). Similarly, our replay-augmented fine-tuned model, \(\mathcal{M}_{\mathrm{f.t}}^{r.p.}\), is also fine-tuned from the same base model \(\mathcal{M}_{0}\). The table below summarizes the notation used for clarity.

\begin{table}[H]
\centering
\label{tab:model_notation}
\renewcommand{\arraystretch}{1.4}
\begin{tabular}{ll}
\toprule
\textbf{Notation} & \textbf{Description} \\
\midrule
\(\mathcal{M}_{0}\) & Base model:SVD \\
\(\mathcal{M}_{\mathrm{f.t}}\) & Vista~\cite{gao2024vista}: Fine-tuned from \(\mathcal{M}_{0}\) (SVD) \\
\(\mathcal{M}_{\mathrm{f.t}}^{r.p.}\) & Our fine-tuned model from \(\mathcal{M}_{0}\), enhanced with offline replay \\
\bottomrule
\end{tabular}
\end{table}

\section{Additional Experiments and Discussion}
\subsection{Evaluation on \(\mathcal{M}_{\mathrm{f.t}}^{r.p.}\)}

See \Cref{sec:imp_rp} for implementation details. Note that \(\mathcal{M}_{\mathrm{f.t}}^{r.p.}\) is not intended as a novel approach or architectural contribution. Rather, the purpose of this replay-based model is to verify that the degradation observed during fine-tuning resembles catastrophic forgetting, as discussed in \Cref{sec:forgetting}, and that such effects can be mitigated using standard continual learning techniques.
Our results show that \(\mathcal{M}_{\mathrm{f.t}}^{r.p.}\) slightly improves visual quality (\Cref{tab:fid_fvd}) while maintaining similar performance on spatial and temporal metrics (\Cref{tab:duration_stats,tab:line_n}) and identical performance on the traffic sign evaluation task. Specifically, the model achieves 5\% accuracy in recognizing traffic signs and 30\% accuracy in correctly stopping at red lights, matching the baseline model without any performance degradation, offer a balanced alternative by preserving spatial accuracy while maintaining strong visual quality.

\begin{wrapfigure}{r}{0.3\textwidth}
\vspace{-4mm}
    \centering
    \captionsetup{skip=5pt}
    \includegraphics[width=\linewidth]{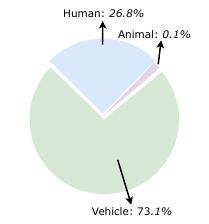}
    \caption{}
    \label{fig:category}
    \vspace{-3mm}
\end{wrapfigure}
\subsection{Experiment on Different SCA Categories}

\begin{figure}[t]
  \centering
  \begin{minipage}{0.32\linewidth}
    \centering
    \includegraphics[width=1\linewidth]{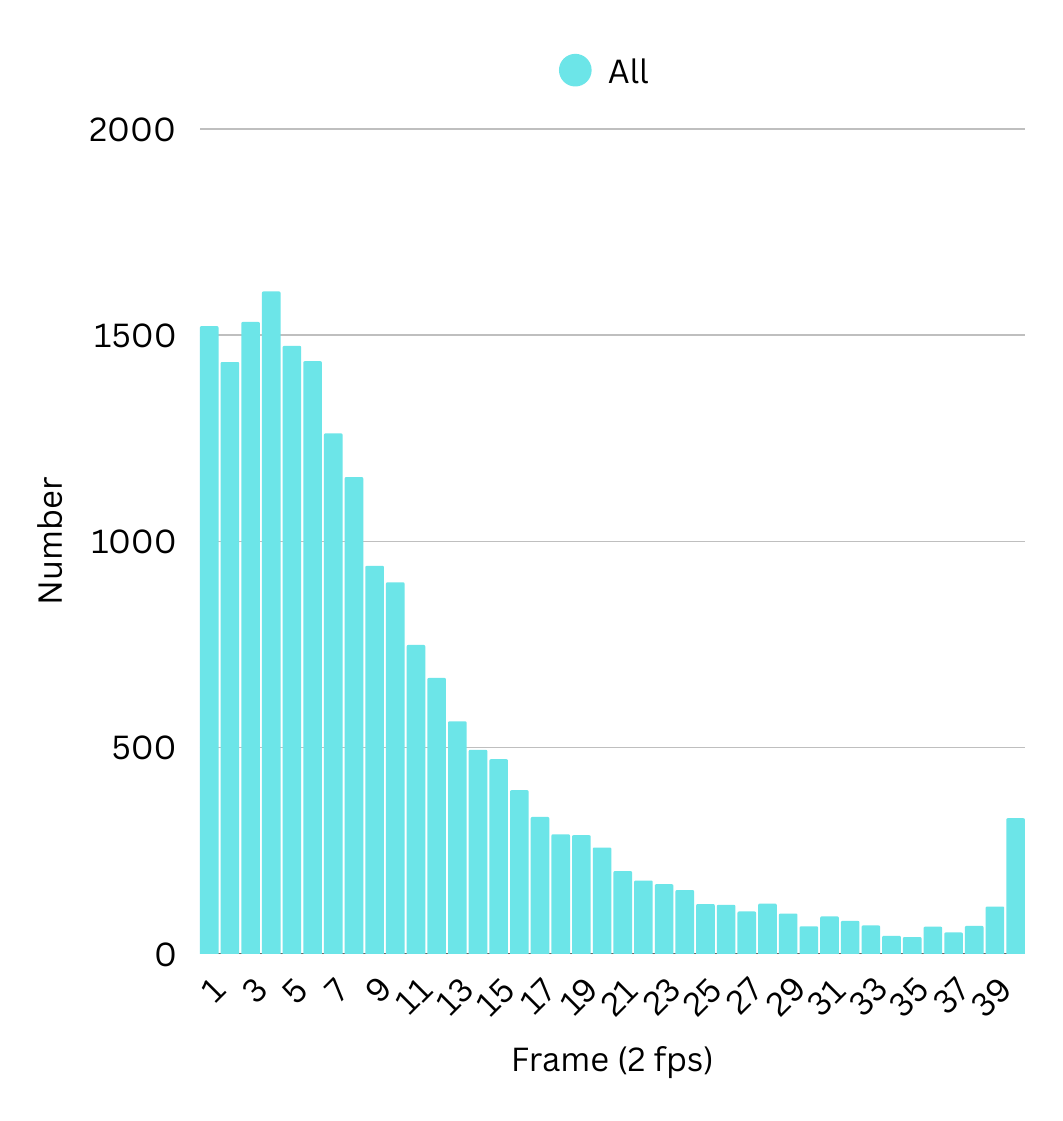}
    \label{fig:duration_all}
  \end{minipage}
  \hfill
  \begin{minipage}{0.32\linewidth}
    \centering
    \includegraphics[width=1\linewidth]{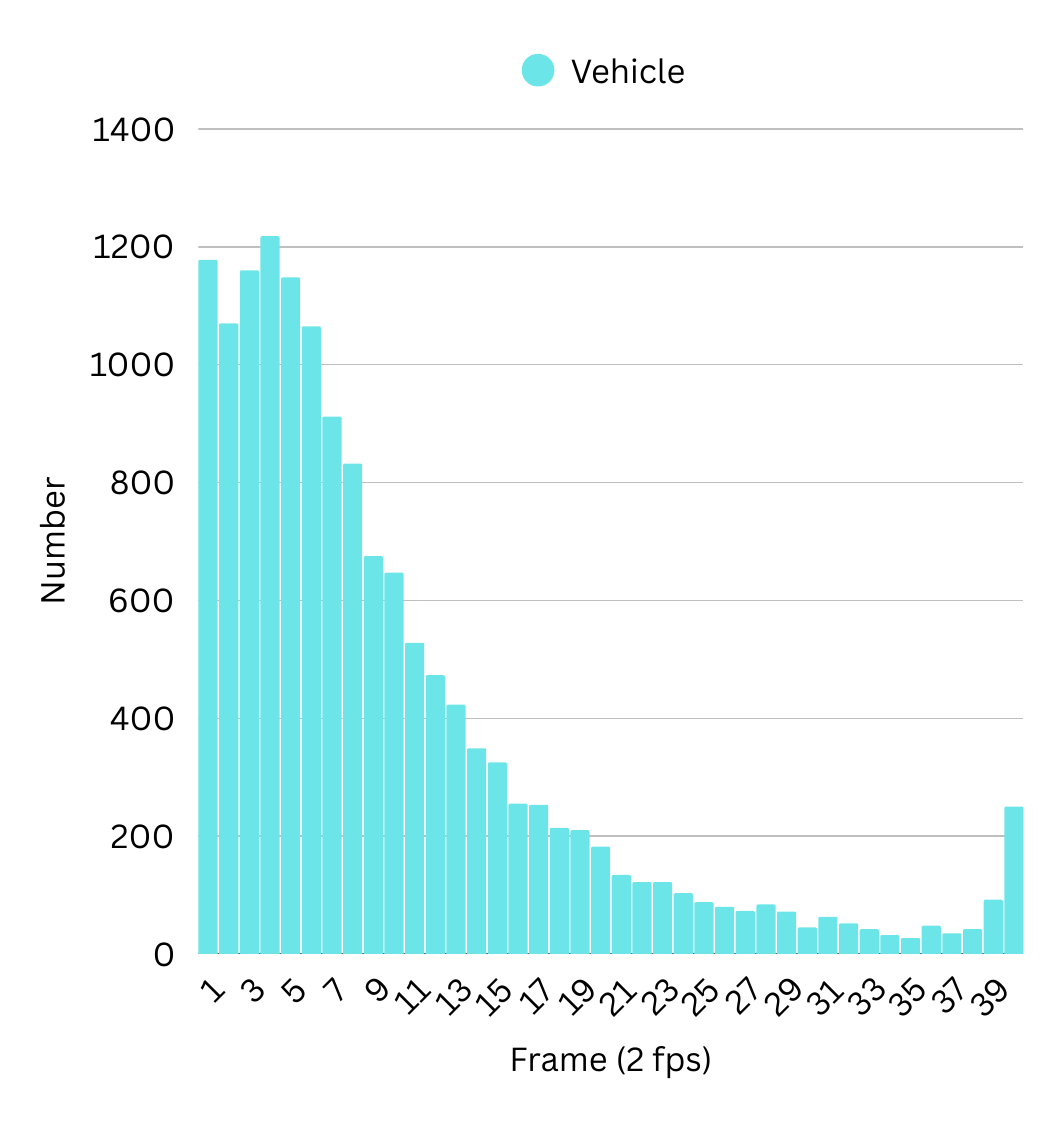}
    \label{tab:duration_vehicle}
  \end{minipage}
  \begin{minipage}{0.32\linewidth}
    \centering
    \includegraphics[width=1\linewidth]{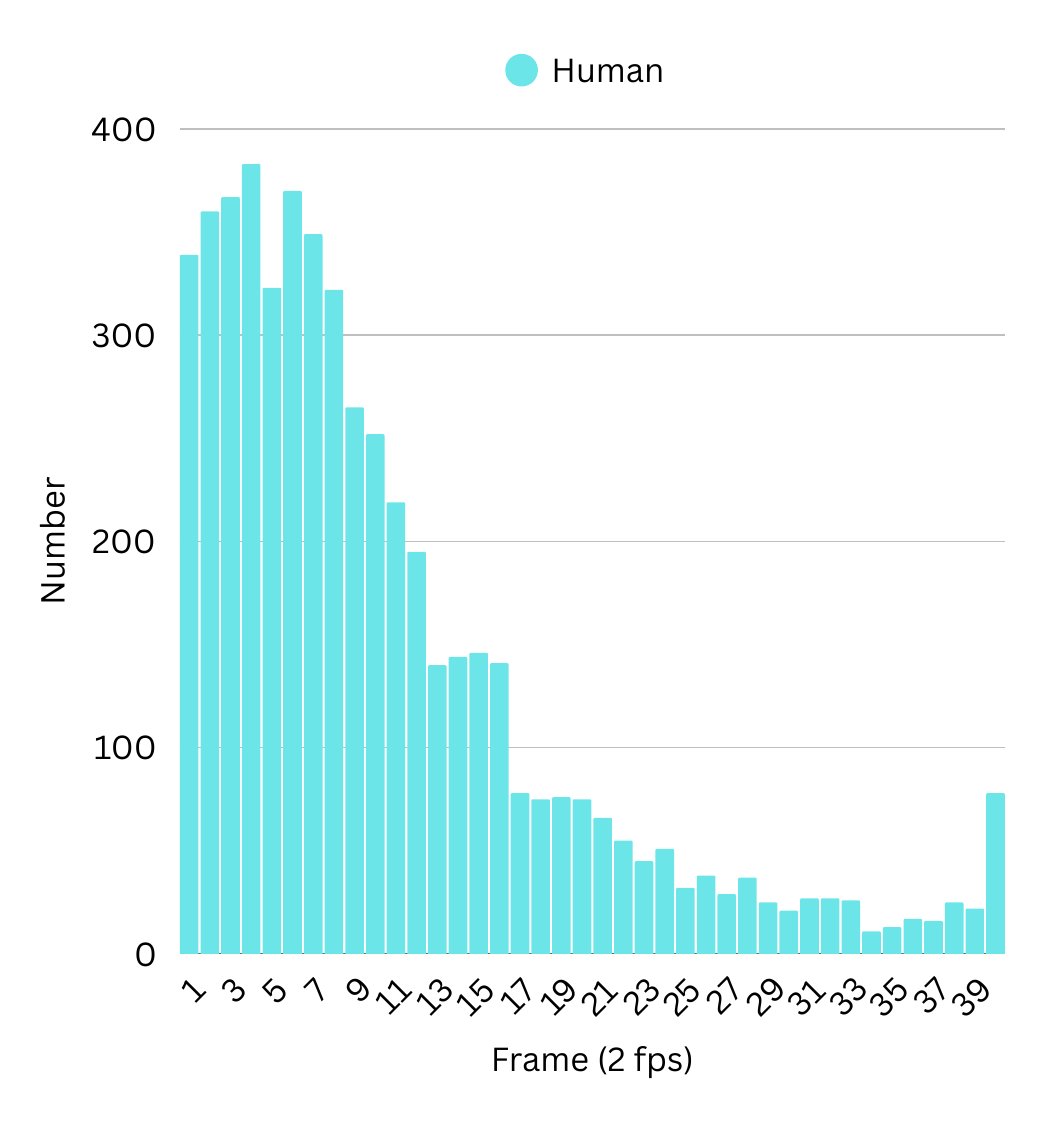}
    \label{tab:duration_human}
  \end{minipage}
  \caption{Distribution of SCAs presence duration in the nuScenes dataset, shown for all objects (left), vehicles (middle), and humans (right). The x-axis represents frame number at 2 FPS, corresponding to time duration in the scene. Animal instances are excluded due to insufficient sample count. Most humans and vehicles appear for less than 10 seconds. The peak at the final frame suggests either the ego vehicle is stationary or following a leading vehicle without overtaking.}
  \label{tab:duration}
\end{figure}

In nuScenes, we focus on three main types of SCAs: humans, vehicles, and animals. However, animal instances account for only about 0.1\% of all SCAs, so our analysis primarily focuses on humans and vehicles (see \Cref{fig:category}). As shown in \Cref{tab:duration}, approximately 72.9\% of humans and 68.9\% of vehicles remain visible in the scene for at least 2.5 seconds. This indicates that under the commonly used evaluation window~\cite{gao2024vista, hassan2024gem} (2.5 seconds), the majority of SCAs remain present throughout the evaluation period.

In \Cref{tab:category_pixel_diff}, we analyze the pixel-wise difference between predicted and ground-truth object masks, using SAM2 to track the same SCA across frames. For each frame index, we compute the average difference in mask area between the generated and ground-truth frames. Ideally, the difference should be close to zero if the predicted scene preserves object boundaries accurately.
The results indicate that the fine-tuned model \(\mathcal{M}_{\text{f.t.}}\) consistently reduces mask differences compared to the base model \(\mathcal{M}_0\). This confirms that fine-tuning helps preserve object structure and prevent scene collapse, an issue frequently observed in the baseline model, where blurry object boundaries often result in substantial discrepancies in SAM2, generated masks.

\begin{table}[t]
\centering
\caption{
Average pixel-wise difference and standard deviation between SAM2-predicted masks on generated frames and the corresponding ground-truth frames, computed per frame index for human and vehicle. A lower average difference indicates better spatial consistency and preservation of object structure. The fine-tuned model \(\mathcal{M}_{\text{f.t.}}\) consistently shows reduced discrepancies compared to the base model \(\mathcal{M}_0\), especially in later frames, highlighting its effectiveness in maintaining object boundaries and preventing scene collapse.
}

\label{tab:category_pixel_diff}
\scriptsize
\renewcommand{\arraystretch}{1.4}
\begin{tabular}{l|cccc|cccc}

\toprule
\multirow{2}{*}{Frame ID} & \multicolumn{2}{c}{\(\mathcal{M}_{0}\) (Human)} & \multicolumn{2}{c}{\(\mathcal{M}_{0}\) (Vehicle)} & \multicolumn{2}{c}{\(\mathcal{M}_{\mathrm{f.t}}\) (Human)} & \multicolumn{2}{c}{\(\mathcal{M}_{\mathrm{f.t}}\) (Vehicle)} \\
\cmidrule(r){2-3}
\cmidrule(r){4-5}
\cmidrule(r){6-7}
\cmidrule(r){8-9}
 & Avg Diff.(px) & std & Avg Diff.(px) & std & Avg Diff.(px) & std & Avg Diff.(px) & std \\
\midrule
2 & -30.3 & 483.4 & -87.0 & 1208.2 & -11.5 & 279.4 & -41.6 & 825.1 \\
3 & -7.0 & 526.4 & -53.0 & 1724.9 & -14.1 & 375.4 & -38.5 & 1345.5 \\
4 & 32.0 & 579.4 & -125.9 & 2759.6 & 14.6 & 404.7 & -24.1 & 1700.6 \\
5 & 63.5 & 647.8 & -51.2 & 2971.3 & 25.8 & 545.7 & -13.1 & 2277.0 \\
6 & 128.3 & 785.5 & -24.3 & 3764.6 & 79.8 & 688.9 & 34.0 & 2839.5 \\
7 & 116.1 & 883.1 & -12.3 & 4179.5 & 77.6 & 839.0 & 53.3 & 3189.5 \\
8 & 159.3 & 1040.7 & -21.7 & 4889.9 & 115.0 & 979.1 & 36.7 & 3467.2 \\
9 & 218.2 & 1132.4 & 28.2 & 4534.6 & 138.9 & 1101.4 & 59.0 & 4143.1 \\
10 & 207.3 & 1221.0 & 87.6 & 4944.0 & 130.0 & 1159.3 & 84.3 & 4970.2 \\
11 & 237.9 & 1292.6 & 128.8 & 5474.3 & 143.1 & 1314.6 & 96.5 & 5328.0 \\
12 & 162.3 & 1130.9 & 202.0 & 6090.4 & 89.3 & 1155.0 & 90.1 & 5414.2 \\
13 & 152.9 & 1222.2 & 224.8 & 6658.2 & 64.0 & 1140.1 & 102.2 & 5791.1 \\
14 & 137.0 & 1419.3 & 217.7 & 7141.4 & 55.9 & 1337.0 & 80.8 & 6302.9 \\
15 & 179.1 & 1531.3 & 158.6 & 7590.0 & 80.9 & 1416.1 & 4.25 & 6942.2 \\
16 & 257.2 & 1578.8 & 267.7 & 7714.4 & 181.3 & 1911.9 & 86.7 & 6952.3 \\
17 & 352.2 & 1775.5 & 313.7 & 8072.5 & 225.0 & 1875.6 & 126.7 & 7423.1 \\
18 & 218.8 & 2060.2 & 428.9 & 8435.2 & 104.0 & 1777.2 & 332.5 & 7782.3 \\
19 & 147.0 & 2232.7 & 373.6 & 8944.3 & 101.3 & 1933.9 & 269.5 & 8232.0 \\
20 & 215.6 & 2408.3 & 350.1 & 9458.3 & 204.9 & 2056.4 & 263.9 & 8515.4 \\
21 & 234.3 & 2703.6 & 337.8 & 10154.8 & 343.2 & 2088.0 & 231.3 & 9037.0 \\
22 & 318.1 & 2911.7 & 386.3 & 10652.7 & 453.5 & 2315.7 & 247.5 & 9597.3 \\
23 & 269.5 & 2589.7 & 377.1 & 11067.2 & 347.1 & 2112.5 & 288.5 & 10255.7 \\
24 & 266.4 & 2409.7 & 246.0 & 11749.7 & 245.2 & 1782.5 & 291.8 & 10657.6 \\
25 & 293.7 & 2166.2 & 207.9 & 12424.6 & 250.0 & 1823.0 & 193.3 & 11071.7 \\
\bottomrule
\end{tabular}
\end{table}

\subsection{AI Judge Evaluation}

To address the limitations of human evaluation, such as high annotation effort and poor reproducibility, we utilize GPT-4o as an automated judge. The prompt design is described in \Cref{sec:gpt_judge_prompt}. We compare both the fine-tuned  model (Vista) and our replay-augmented fine-tuned model \(\mathcal{M}_{\mathrm{f.t}}^{r.p.}\) against the baseline model \(\mathcal{M}_{0}\) (SVD).
Among 1{,}000 samples, GPT-4o judged the fine-tuned model as better than the baseline in 74\% of cases, while the baseline was preferred in 26\%. For \(\mathcal{M}_{\mathrm{f.t}}^{r.p.}\), 69\% of comparisons favored our model, with 31\% favoring the baseline. These results support the observed improvements in visual quality introduced by fine-tuning and replay.
We emphasize that AI-based evaluation is intended as an efficient and reproducible supplement to human judgment. However, some degree of bias in the judgment model remains inevitable.

\section{Experiment Detail}

\subsection{Implementation of \(\mathcal{M}_{\mathrm{f.t}}^{r.p.}\)}
\label{sec:imp_rp}

We implement \(\mathcal{M}_{\mathrm{f.t}}^{r.p.}\) as a fine-tuned version of the base model \texttt{SVD\_xt}~\cite{blattmann2023stable}, enhanced with offline replay to mitigate degradation in dynamic modeling. While the original SVD\_xt model is trained on large-scale general-domain video datasets, it is not specifically exposed to EgoExo4D~\cite{grauman2024ego}. In our setup, we use EgoExo4D as a proxy for general-domain diversity during replay, sampling a fixed set of clips that span a wide range of activities such as sports, cooking, and other egocentric scenarios. All replay clips are preprocessed at 10 FPS.

During training, we adopt a per-batch offline replay strategy with a 1:1 sampling ratio between EgoExo4D and nuScenes~\cite{nuscenes}. Each batch contains an equal number of clips from both datasets. This ensures the model is exposed to temporally diverse motion patterns alongside structured driving scenes, helping to preserve its capacity to model dynamic interactions. The replay buffer is fixed and remains unchanged throughout training, making this an offline replay approach.

\subsection{Bounding Box Projection for SAM2 Prompting}

Since the bounding boxes provided in the nuScenes dataset are defined in global coordinates, we apply a series of coordinate transformations to project each 3D bounding box into the image plane before using it as a prompt for SAM2. The process involves the following steps: (1) transforming the box from the global frame to the ego vehicle frame using the ego pose, (2) transforming from the ego frame to the camera frame using the calibrated sensor parameters, and (3) projecting the resulting 3D box onto the image plane using the camera intrinsic matrix. The final 2D bounding box, defined by the projected corner coordinates, serves as input to SAM2, as shown in \Cref{algo:bbox_3d_2_2d}.

\begin{algorithm}[H]
\caption{Project Global 3D Box to 2D Image Coordinates}
\begin{algorithmic}[1]
\Require Global 3D box $\mathcal{B}$, ego pose $E$, camera calibration $C$, camera intrinsics $K$
\Ensure 2D bounding box $(x_\text{min}, y_\text{min}, x_\text{max}, y_\text{max})$
\State Translate $\mathcal{B}$ by $-E_\text{translation}$ and rotate by $E_\text{rotation}^{-1}$ to convert from global to ego frame
\State Translate $\mathcal{B}$ by $-C_\text{translation}$ and rotate by $C_\text{rotation}^{-1}$ to convert from ego to camera frame
\State Compute 3D corners of $\mathcal{B}$ and project them to 2D using intrinsics $K$
\State Extract projected 2D corner coordinates
\State Compute bounding box: 
\[
x_\text{min} = \min(x), \quad y_\text{min} = \min(y), \quad x_\text{max} = \max(x), \quad y_\text{max} = \max(y)
\]
\State \Return $(x_\text{min}, y_\text{min}, x_\text{max}, y_\text{max})$
\end{algorithmic}
\label{algo:bbox_3d_2_2d}
\end{algorithm}

\subsection{Calculate FID/FVD}

Due to differences in resolution and aspect ratio across ground-truth videos and outputs from various generative models, the computation of FID and FVD requires consistent preprocessing to ensure fair comparison. We follow the preprocessing protocol used in Vista~\cite{gao2024vista}, as shown in \Cref{algo:resize}. All frames are first center-cropped to match a fixed aspect ratio before being resized to \(256 \times 448\), ensuring compatibility with image-based metrics like FID. For FVD computation, the cropped and resized frames undergo an additional resizing step to \(224 \times 224\) using bilinear interpolation. This pipeline ensures that both FID and FVD are computed on inputs that share the same aspect ratio and resolution standards.

\begin{algorithm}[H]
\caption{Preprocessing for FID and FVD Computation}
\begin{algorithmic}[1]
\Procedure{CropAndResize}{$\mathcal{I}, w=448, h=256$}
    \State Load image \(\mathcal{I}\), get original dimensions \((W, H)\)
    \If{\(W / H > w / h\)} 
        \State Crop width to match aspect ratio: center-crop width to \(w'\)
    \ElsIf{\(W / H < w / h\)}
        \State Crop height to match aspect ratio: center-crop height to \(h'\)
    \EndIf
    \State Resize image to \((w, h)\) using Lanczos resampling
    \State \Return Preprocessed image
\EndProcedure

\Procedure{PreprocessForFID}{$\mathcal{I}$}
    \State \Return \Call{CropAndResize}{$\mathcal{I}, 448, 256$}
\EndProcedure

\Procedure{PreprocessForFVD}{$\mathcal{V}$}
    \For{each frame \(\mathcal{I} \in \mathcal{V}\)}
        \State \(\mathcal{I'} \gets \Call{CropAndResize}{\mathcal{I}, 448, 256}\)
        \State Resize \(\mathcal{I'}\) to \((224, 224)\) using bilinear interpolation
    \EndFor
    \State \Return Sequence of resized frames
\EndProcedure
\end{algorithmic}
\label{algo:resize}
\end{algorithm}

\begin{figure}[H]
\captionsetup{skip=3pt}
  \centering
   \includegraphics[width=0.9\linewidth]{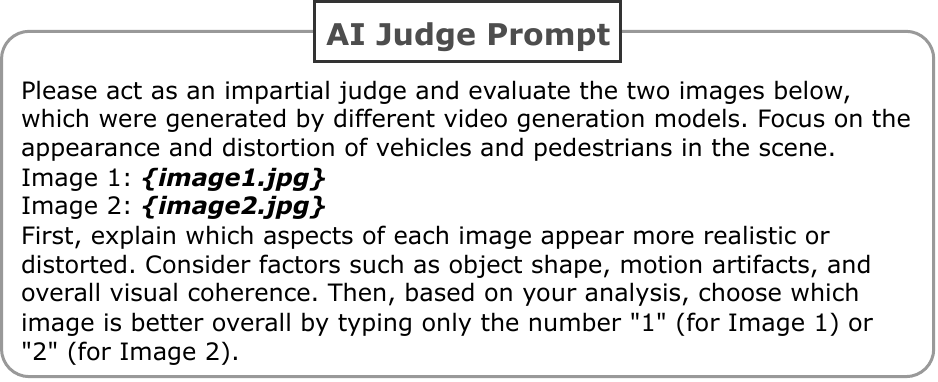}
   \caption{
}
   \label{fig:gpt_prompt}
   \vspace{-3mm}
\end{figure}
\subsection{AI as Judge}
\label{sec:gpt_judge_prompt}
Evaluating visual quality is inherently challenging and often subjective. While conventional metrics such as FID and FVD can capture certain aspects of quality, their results do not always align with human judgment. Consequently, human evaluation has become increasingly popular. However, it also suffers from limitations, including limited reviewer availability and individual bias, making the results difficult to reproduce or compare across studies. To address these issues, recent work has explored the use of AI assistants, such as GPT-4o, as automated judges. Although AI-based evaluation can also exhibit bias and should always be complemented by other metrics, existing studies~\cite{zheng2023judgingllmasajudgemtbenchchatbot,chiang2023can, chang20243d, chang2024mikasa} have shown a high level of agreement between large language model evaluations and human judgments. In \Cref{fig:gpt_prompt}, we provide the template used for GPT-4o-based evaluation.


\end{document}